\documentclass[11pt,a4paper]{article}

\usepackage[utf8]{inputenc}
\usepackage{times}
\usepackage{geometry}
\usepackage{graphicx}
\usepackage{hyperref}
\usepackage{booktabs}
\usepackage{multirow}
\usepackage{caption}
\usepackage{subcaption}
\usepackage{tikz}
\usetikzlibrary{arrows.meta, positioning, fit, backgrounds, calc}
\usepackage[table]{xcolor}
\usepackage{xcolor}
\usepackage{array}
\usepackage{tabularx}
\usepackage[section]{placeins}

\title{\textbf{UniK: Universal Knowledge Perception\\for Digital and Physical AI}}

\author{
Nirmit Desai\thanks{Corresponding author: \texttt{nirmit.desai@aintropy.ai}} \quad
Kunal Sawarkar \quad
Aditya Mahakali \quad
Dongkon Lee \quad
Kevin Park \quad
Eric Song \\[4pt]
AIntropy AI
}

\date{Sep 2026}

\begin{document}

\maketitle

\begin{abstract}
Two transformative classes of AI systems are reshaping how organizations operate: \textit{digital AI}, which reasons over enterprise knowledge to power chatbots and agent workflows; and \textit{physical AI}, which learns to control robots and autonomous systems from video, gameplay, and sensor telemetry. Both face the same foundational bottleneck: raw knowledge at scale, spanning heterogeneous modalities, locked in private corpora that existing AI infrastructure cannot access reliably or efficiently. We propose \textit{Universal Knowledge Perception (UniK)} as a common platform for both classes, covering the full knowledge lifecycle (ingestion, enrichment, indexing, retrieval, and continuous evaluation) across modalities from rich text and video to molecular data and sensor telemetry. We present UniK, built on Polymath Retrieval (multi-index fusion over automatically enriched indices) with no task-specific fine-tuning. Across five digital AI domains (medical literature, open-domain QA, chemistry, legal video proceedings, and government open data) UniK combined with an open-source 70-billion-parameter model consistently matches or outperforms frontier proprietary LLMs that are orders of magnitude larger: 76\% RAG accuracy on government data versus 47\% for GPT-5; 77.9\% on medical QA without fine-tuning; topping all open-source chemistry pipelines. We show that the same infrastructure directly addresses the data curation, indexing, and retrieval challenges facing physical AI world model training, where the knowledge problem is harder but structurally identical.
\end{abstract}

\section{Introduction}

Two distinct classes of AI systems are converging toward a common knowledge perception problem. \textit{Digital AI} (large language models powering chatbots and agent workflows) must reason accurately over private enterprise knowledge bases that were never part of their training data. \textit{Physical AI} (robots, autonomous vehicles, and embodied agents) must learn world models from video, gameplay, and sensor telemetry collected at industrial scale. Both classes share the same core challenge: the knowledge they need to function is heterogeneous, domain-specific, multi-modal, and far too large to fit in any model's context window. 

In both cases, the bottleneck is not model intelligence but \textit{knowledge perception}: the ability to reliably access and surface the right knowledge at the right time. Perception is distinguished from the \textit{understanding} or reasoning layer, which is the province of the language models or world models that consume retrieved knowledge. LeCun~\cite{lecun2022jepa} draws a related distinction: perception extracts and structures information from the world, while understanding requires a predictive world model capable of reasoning and planning. This paper focuses on the perceptual substrate that any reasoning system depends on. A stronger retrieval and enrichment platform makes any model more capable on knowledge-intensive tasks, regardless of the model's intrinsic reasoning ability. Through experimental results across domains, we show that the model's understanding and reasoning is no longer the limiting factor, the perception layer is.

\begin{figure}[htbp]
\centering
\setlength{\fboxsep}{0pt}\setlength{\fboxrule}{0.3pt}
\begin{subfigure}[t]{0.235\textwidth}
  \fbox{\includegraphics[width=\linewidth]{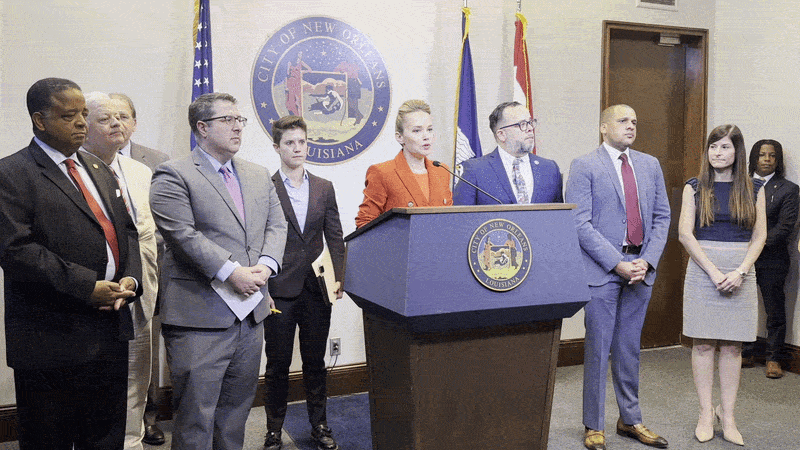}}
  \vspace{2pt}
  \par\scriptsize\itshape\sffamily``What was the council's decision on mobility and street services at this session?''
\end{subfigure}\hfill
\begin{subfigure}[t]{0.235\textwidth}
  \fbox{\includegraphics[width=\linewidth]{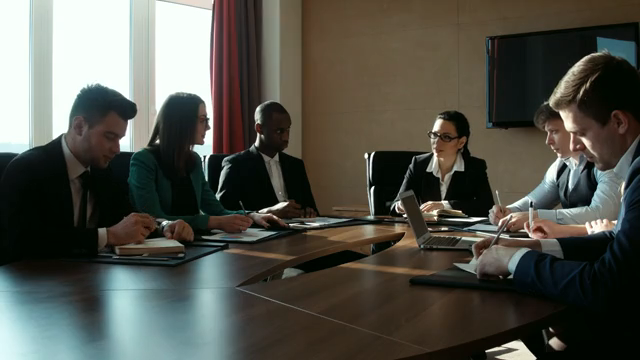}}
  \vspace{2pt}
  \par\scriptsize\itshape\sffamily``What are the top 3 features Customer X needs in our product to choose us over competitors?''
\end{subfigure}\hfill
\begin{subfigure}[t]{0.235\textwidth}
  \fbox{\includegraphics[width=\linewidth]{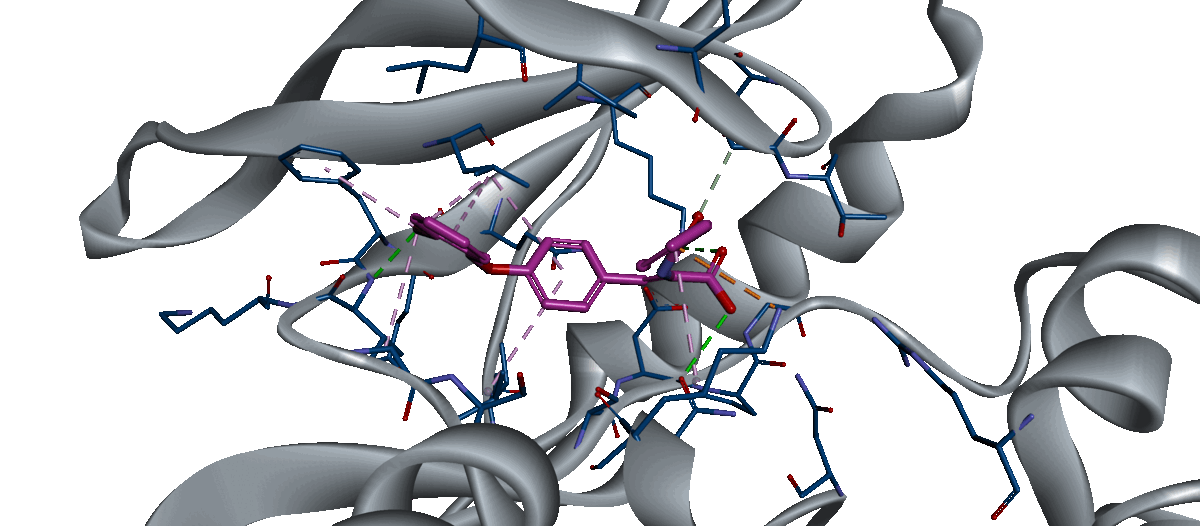}}
  \vspace{2pt}
  \par\scriptsize\itshape\sffamily``Find all drug interaction risks across 100M clinical records that current labeling doesn't cover.''
\end{subfigure}\hfill
\begin{subfigure}[t]{0.235\textwidth}
  \fbox{\includegraphics[width=\linewidth]{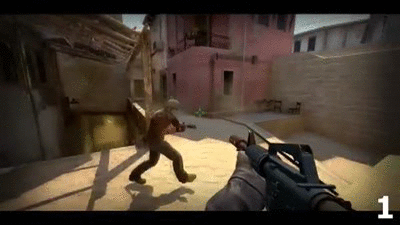}}
  \vspace{2pt}
  \par\scriptsize\itshape\sffamily``Retrieve episodes where the agent executes the objective under time pressure.''
\end{subfigure}

\vspace{8pt}

\begin{subfigure}[t]{0.235\textwidth}
  \fbox{\includegraphics[width=\linewidth]{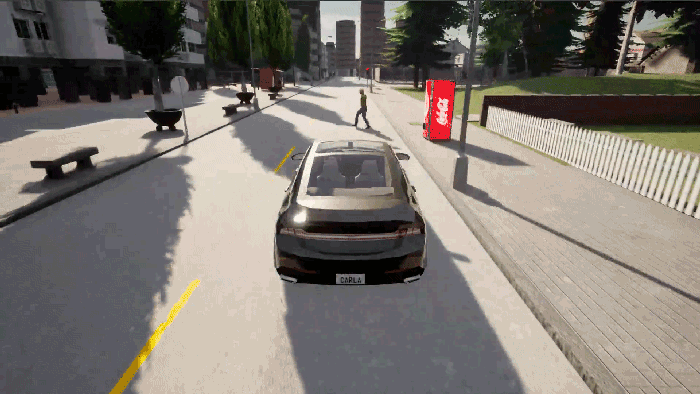}}
  \vspace{2pt}
  \par\scriptsize\itshape\sffamily``Apply physics constraints and curate these driving simulation episodes for diversity across road conditions, weather, and time of day.''
\end{subfigure}\hfill
\begin{subfigure}[t]{0.235\textwidth}
  \fbox{\includegraphics[width=\linewidth]{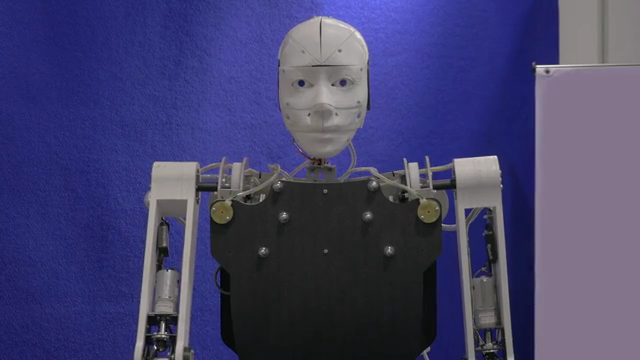}}
  \vspace{2pt}
  \par\scriptsize\itshape\sffamily``Which movement sequences from this demo exhibit the highest balance error?''
\end{subfigure}\hfill
\begin{subfigure}[t]{0.235\textwidth}
  \fbox{\includegraphics[width=\linewidth]{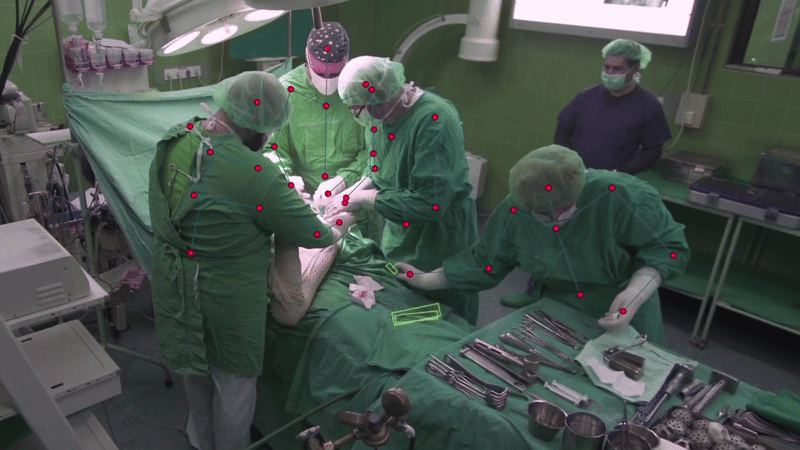}}
  \vspace{2pt}
  \par\scriptsize\itshape\sffamily``Across 800 surgical videos, find every case where a diabetic patient had anesthesia complications and cross-reference with EHR data.''
\end{subfigure}\hfill
\begin{subfigure}[t]{0.235\textwidth}
  \fbox{\includegraphics[width=\linewidth]{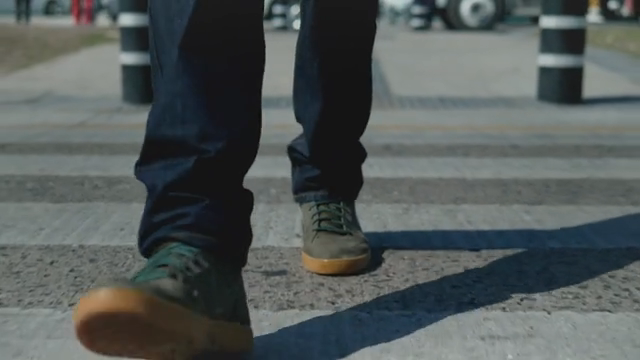}}
  \vspace{2pt}
  \par\scriptsize\itshape\sffamily``Which patrol footage segments from the past 7 days match the incident described in report \#4821?''
\end{subfigure}

\caption{The knowledge perception problem across eight enterprise and AI domains. Row~1 (left to right): city council proceedings, corporate sales meeting, pharmaceutical drug interaction records, and gameplay training data. Row~2: autonomous driving simulation, humanoid robot demonstration, surgical video with EHR cross-reference, and body-worn camera footage. In every case, relevant knowledge is locked in unindexed video streams, unstructured documents, or heterogeneous sensor and telemetry logs. The bottleneck is knowledge perception, not model capability.}
\label{fig:knowledge-problem}
\end{figure}

Figure~\ref{fig:knowledge-problem} makes this concrete. A paralegal reviewing a zoning dispute must locate specific testimony across hundreds of hours of city council and legislative video, with every word locked in the video stream and invisible to any search tool. A robotics engineer curating training data for a manipulation policy needs the subset of collected gameplay footage that demonstrates a particular skill sequence; no semantic index exists to retrieve it. Both problems fail at the perception layer, not the model layer.

For digital AI, this manifests as the \textit{great knowledge divide}. More than 95\% of enterprise AI pilots fail to deliver return on investment~\cite{mit2025genaidivide}, not because the underlying models are incapable, but because the knowledge that drives enterprise value (regulations, clinical notes, legal proceedings, proprietary research, government datasets) lives behind firewalls in formats and modalities that public pretraining never touched. Two paradigms have been proposed to bridge this gap. Fine-tuning adapts an LLM to domain knowledge directly, but causes catastrophic forgetting~\cite{siriwardhana-etal-2023-improving}, is sensitive to training hyperparameters, and must be repeated as knowledge evolves. Retrieval-Augmented Generation (RAG) is architecturally more sound: it retrieves context at inference time rather than baking knowledge into weights, but it degrades at scale: dense retrieval accuracy falls as corpus size grows~\cite{reimers2021cursedenselowdimensional, weller2026-embeddingbasedretrieval}, off-the-shelf embeddings fail in specialized domains~\cite{barnett2024sevenfailurepoints}, and context window limits bound how much retrieved knowledge an LLM can actually use~\cite{hsieh2024rulerwhatsrealcontext, du2025contextlengthhurtsllm}.

The mathematical foundations of this failure at scale are well understood and have been studied for decades~\cite{thakur2021beir}. Dense retrieval maps both queries and documents to vectors in a shared embedding space and relies on geometric proximity as a proxy for semantic relevance. This assumption degrades structurally as corpus size grows through two mechanisms. First, in high-dimensional spaces the \textit{hubness problem} emerges: a small fraction of document vectors become the nearest neighbors for a disproportionately large fraction of queries regardless of semantic content, biasing retrieval toward a handful of ``hub'' documents and away from the long-tail items most relevant to specialized queries~\cite{radovanovic2010hubs, reimers2021cursedenselowdimensional}. Second, general-purpose embedding models collapse domain-specific vocabulary: a SMILES string and its systematic chemical name, a medical acronym and the condition it abbreviates, or a legislation reference and its colloquial title may land far apart in an embedding space trained on general web text, even though they refer to the same entity~\cite{barnett2024sevenfailurepoints, weller2026-embeddingbasedretrieval}. The practical consequence is striking: BM25, introduced in the 1990s, remains competitive with neural retrieval on heterogeneous zero-shot benchmarks today~\cite{thakur2021beir}, while LLMs have progressed from near-random to near-human on language understanding tasks within a single decade. Retrieval, not model intelligence, is the binding constraint for knowledge-intensive applications, a constraint that manifests across agent memory, enterprise chatbots, scientific literature mining, and physical AI data pipelines alike.

For physical AI, the same knowledge perception problem appears in a harder form, under physical and real-time constraints that text retrieval never faces (Section~\ref{sec:open_challenges}). A world model must be trained on episodes that are semantically relevant (similar tasks and object configurations), physically valid (successful manipulations or informative failures), distribution-covering (diverse enough to prevent mode collapse), and sequenced for curriculum learning. These properties cannot be specified by hand-labeled metadata at the scale of millions of video episodes or billions of telemetry samples. They require the same operations that digital AI demands of its knowledge infrastructure: automated enrichment, multi-signal indexing, and efficient cross-index retrieval under task-specific queries---formalized in Section~\ref{sec:platform} as Polymath Enrichment, Polymath Indexing, and Polymath Retrieval. The difference is modality (actions, trajectories, and sensor readings instead of text), but the knowledge perception problem is structurally identical.

We propose \textit{Universal Knowledge Perception (UniK)} as a foundational platform that addresses both classes of AI jointly, spanning the full knowledge lifecycle: discovery, ingestion, enrichment, indexing, retrieval, and continuous evaluation. We argue that these six operations constitute a common platform for digital and physical AI alike, and that investing in this platform rather than chasing larger models or domain-specific adapters is the most tractable path to production-ready AI across the enterprise.

We present UniK and validate it across five digital AI domains spanning text, video, chemical, and structured data modalities. Strong knowledge retrieval compensates for large differences in model scale: UniK combined with a the opensource Llama3 70B instruct model (multiple versions tested across 3.1 and 3.3) consistently matches or outperforms frontier proprietary models across domains where grounded knowledge is essential. For physical AI, we characterize the knowledge infrastructure challenge across three use cases (world model training from gameplay and teleoperation, visual encoder pretraining from egocentric video, and operational intelligence from deployed robot fleets), and show that the same UniK architecture applies directly, with modality-specific enrichment and indexing as the primary points of differentiation.

\section{Universal Knowledge Perception}

\subsection{UniK Platform Requirements}

We define Universal Knowledge Perception as the capability of an AI system to reliably access and utilize any relevant knowledge at the time it is needed, across the full diversity of knowledge types that digital and physical AI systems encounter. Five requirements must hold simultaneously:

\begin{itemize}
  \item \textbf{Breadth}: operates across modalities (text, video, structured data, molecular/chemical, and sensor/telemetry) without requiring modality-specific pipelines. For digital AI this means a single platform serving medical literature, legal video, and government datasets; for physical AI it means the same platform ingesting robot telemetry, gameplay recordings, and egocentric video.
  \item \textbf{Scale}: maintains high retrieval accuracy as the knowledge corpus grows from thousands to billions of items. Digital AI corpora span millions of documents; physical AI corpora span billions of timestamped sensor samples and thousands of hours of video.
  \item \textbf{Domain generality}: achieves strong performance across domains (medicine, law, chemistry, government, robotics) without task-specific fine-tuning. No domain should require bespoke adapters; the platform should generalize by construction.
  \item \textbf{Currency}: accommodates continuously evolving knowledge without retraining. Enterprise documents are updated daily; robot fleet data streams continuously.
  \item \textbf{Efficiency}: delivers knowledge at latencies compatible with the application: interactive response times for digital AI agents, near-real-time episode retrieval for physical AI training pipelines.
\end{itemize}

Meeting all five requirements simultaneously is what makes UniK challenging to build. Any single existing technique satisfies some but not all: dense retrieval degrades at scale; fine-tuned models go stale; domain-specific adapters do not generalize; keyword search misses semantic meaning. The platform must compose multiple techniques in a synergistic way.

\subsection{The Knowledge Spectrum}

Enterprise knowledge exists on a spectrum of modalities, each with distinct retrieval challenges:

\textbf{Rich text} (reports, papers, regulations, contracts) is the most studied, yet even here domain-specific terminology creates retrieval failures when vocabulary gaps between query and document are large.

\textbf{Video} (legal proceedings, government meetings, training footage, product demonstrations) encodes information in the temporal relationship between visual scenes and speech that text transcription alone cannot capture; accurate retrieval requires multimodal enrichment.

\textbf{Structured data} (government databases, scientific datasets, enterprise records) demands that natural language queries be translated to structured queries while also leveraging the semantic content of records.

\textbf{Molecular and chemical data} combines structured identifiers (SMILES, IUPAC names) with unstructured literature at scales exceeding 100 million documents, spanning tasks from property prediction to reaction synthesis.

\textbf{Sensor and telemetry data} from deployed physical systems (robots, autonomous vehicles, industrial equipment) is high-frequency, causally structured, and presents a fundamentally different retrieval problem than text: relevant episodes must be located by temporal and causal relationship to a target behavior or failure mode, not by semantic similarity to a text query. This modality is the primary substrate for physical AI world model training and operational intelligence, making it as central to UniK as text is for digital AI.

\begin{figure*}[htbp]
\centering
\includegraphics[width=\textwidth]{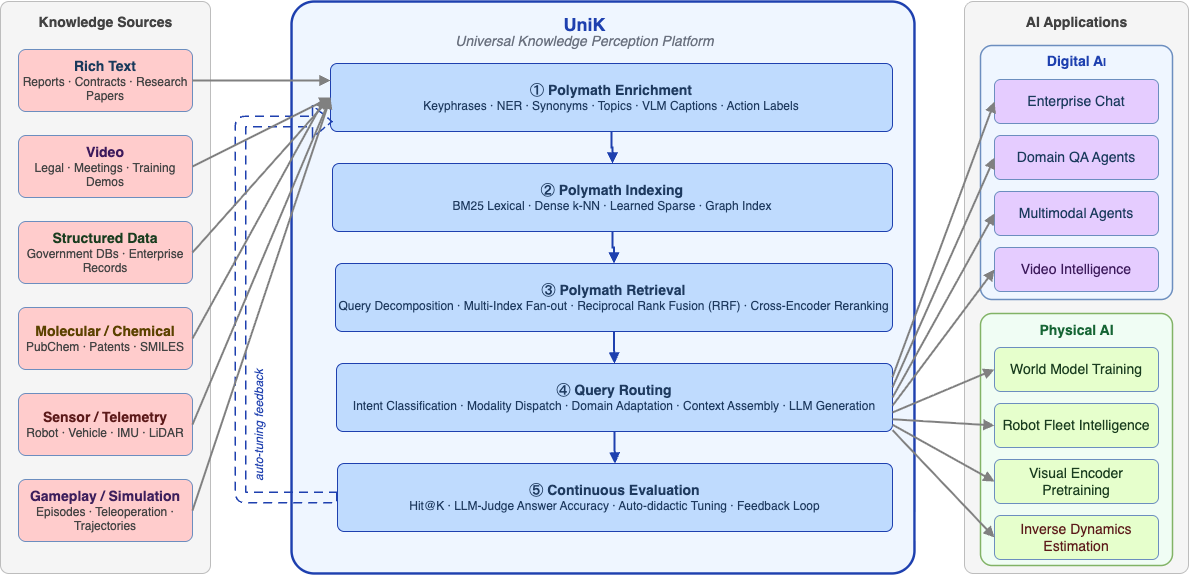}
\caption{The UniK platform: a common knowledge infrastructure for digital and physical AI.
Six heterogeneous knowledge source modalities (left) feed into five core engine operations (center, numbered 1--5):
Polymath Enrichment, Polymath Indexing, Polymath Retrieval, Query Routing, and Continuous Evaluation, with an auto-didactic feedback loop from Continuous Evaluation back to Enrichment.
The same platform serves both digital AI applications (Enterprise Chat, Domain QA Agents, Multimodal Agents, Video Intelligence) and physical AI applications (World Model Training, Robot Fleet Intelligence, Visual Encoder Pretraining, Inverse Dynamics Estimation) without structural modification.}
\label{fig:UniK_overview}
\end{figure*}

\section{The UniK Platform}
\label{sec:platform}

UniK is designed to serve both digital and physical AI with a single unified architecture. For digital AI, it retrieves relevant documents from large enterprise corpora at inference time. For physical AI, it retrieves semantically and physically relevant training episodes from large video and telemetry corpora at data-pipeline time. The operations are identical; the modality adapters differ. Building one platform rather than two separate systems is a deliberate architectural choice: improvements to enrichment, indexing, and fusion compound across every downstream use case simultaneously.

The engine is governed by two design principles: \textit{no task-specific fine-tuning} (the same pipeline generalizes across domains and modalities without per-domain training) and \textit{polymath by default}~\cite{sawarkar2024blended}: no single indexing or retrieval method reliably dominates at scale, so the engine combines lexical, dense, and learned-sparse signals. 

Multi-modality is widely studied, and yet falls short of addressing the challenges identified above. We use the term \textbf{Polymath} to indicate a diversity of techniques, to address not only multi-modal inputs but also distinct domain requirements. Thus, \textbf{Polymath Retrieval} is the engine's query-time fusion operation: a user query is decomposed, routed to one or more index types, and the resulting ranked lists are fused via Reciprocal Rank Fusion (RRF). The companion operation is \textbf{Polymath Indexing}: constructing multi-type BM25, dense k-NN, and learned-sparse indices leveraging \textbf{Polymath Enrichment} to augment the metadata fields: keyphrases, named entities, synonyms, topics, and captions extracted automatically~\cite{sawarkar2025metagen}. Polymath Enrichment, Indexing, and Retrieval are the three operations that distinguish UniK from pipelines that rely on a single retrieval modality or a commodity store.

A fourth distinguishing capability, named here for conceptual completeness and pursued as future work, is \textbf{Auto-didactic Retrieval}: the automatic optimization of the retrieval engine's hyperparameters---query type, embedding dimension, fusion weights, BM25 field boosts, and RRF $k$---to maximize hit@$k$ for a given domain and corpus without manual configuration. Current deployments require domain-specific parameter choices that must be set by practitioners; auto-didactic retrieval learns these from labeled examples or weak supervision signals, analogous to recent work on online hyperparameter tuning for RAG~\cite{autoragHP}. A parallel line of work automates the choice of pipeline components themselves rather than their hyperparameters: AutoRAG~\cite{autorag2024} searches over chunking strategies, retrieval modules, and reranking configurations to find the best-performing pipeline for a given corpus, which we view as complementary to online hyperparameter tuning---one selects the pipeline shape, the other tunes it continuously. We distinguish \textit{auto-configuration} (one-time per-domain setup at ingestion time) from \textit{auto-tuning} (continuous online adaptation as the corpus and query distribution evolve); both are open engineering challenges whose solution would complete the self-improving knowledge perception loop shown in Figure~\ref{fig:UniK_overview}.

\begin{figure*}[htbp]
\centering
\includegraphics[width=\textwidth]{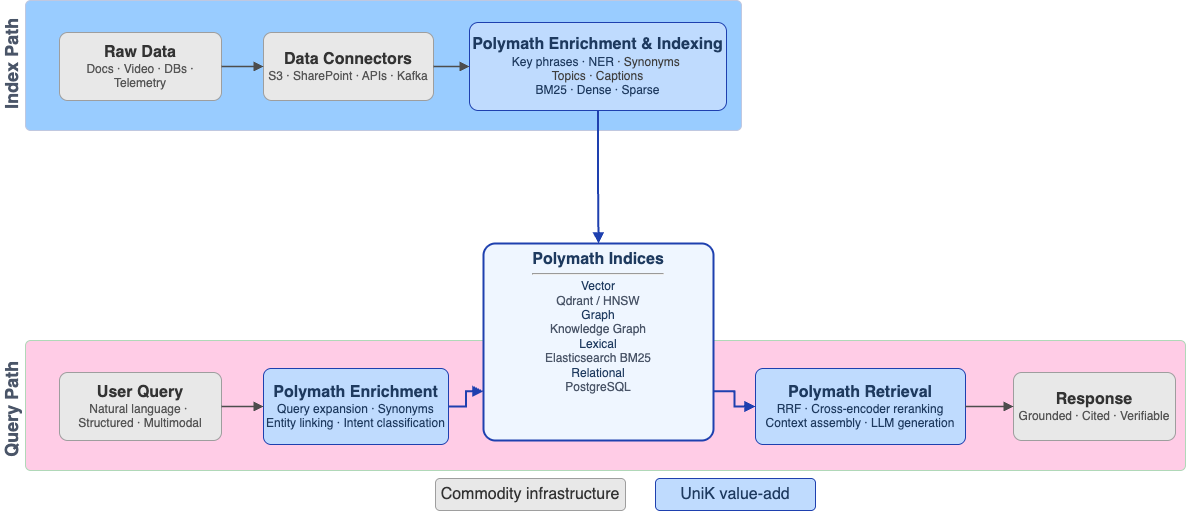}
\caption{End-to-end UniK pipeline.
\textbf{Index path} (top, blue background): raw data flows through Data Connectors into \textit{Polymath Enrichment \& Indexing} (UniK, blue)---which applies KeyBERT, YAKE, spaCy NER, VLMs, and multi-type index construction (BM25, dense, sparse)---and populates the central \textit{Polymath Indices} (vector, graph, lexical, relational).
\textbf{Query path} (bottom, green background): a user query is first enriched by \textit{Polymath Enrichment} (query expansion, entity linking, intent classification), then fed into the Polymath Indices from the left; results exit right into \textit{Fusion \& Ranking} (RRF, cross-encoder reranking, context assembly), which produces the final grounded response.
Commodity components (gray) require no modification across domains; UniK value-add (blue) lies in enrichment, multi-type indexing, and fusion.}
\label{fig:pipeline}
\end{figure*}

Figure~\ref{fig:pipeline} situates these operations in the end-to-end pipeline. The \textit{indexing path} converts raw data into Polymath Indices through enrichment, embedding, and multi-type indexing. The \textit{query path} decomposes an incoming query, routes it through the Polymath Router to one or more index types, and aggregates results via Fusion \& Ranking. Commodity vector, graph, lexical, and relational stores are the underlying infrastructure; UniK adds value at the enrichment step and at the query decomposition, routing, and fusion steps. Neither path requires structural modification when the modality changes from text to video to sensor telemetry; only the enrichment adapters differ.

This architecture explains why UniK extends naturally to physical AI. The indexing path ingests robot telemetry episodes or egocentric video the same way it ingests documents: enrichment extracts action types, object labels, and kinematic features rather than keyphrases and entities. The query path routes a behavior specification or failure pattern through the same Polymath Router and returns the most relevant training episodes by the same fusion and ranking mechanism. Only the enrichment vocabulary differs.

\section{Evaluation Across Domains}

We evaluate the UniK platform across five domains spanning different modalities, corpus scales, and task types. In all cases, the same underlying Polymath Retrieval architecture is used with no domain-specific tuning. The generation model is Llama-3.3-70B-Instruct~\cite{grattafiori2024llama} throughout, allowing us to isolate the contribution of retrieval quality.

Two distinct evaluation types are used in this section. The medical, open-domain, and chemistry experiments (Sections~\ref{sec:text}--\ref{sec:chem}) use established academic benchmarks: PubMedQA~\cite{jin2019pubmedqadataset}, NQ~\cite{47761}, HotpotQA~\cite{yang2018hotpotqa}, SQuAD~\cite{rajpurkar2016squad}, and the ChemRAG suite~\cite{chemrag}, all with standardized ground-truth labels and public baselines. The legal video and government open data experiments (Sections~\ref{sec:legal}--\ref{sec:gov}) use leaderboards contributed by AIntropy AI~\cite{legalvideo_leaderboard, njod_leaderboard}: curated corpora, question sets, and evaluation protocols designed to reflect realistic production query distributions in those domains.

An important observation about this evaluation: all corpora used across all five domains are publicly accessible. The LocalView and Seattle City Council video transcripts~\cite{legalvideo_leaderboard}, the New Jersey government datasets~\cite{njod_leaderboard}, the ChemRAG corpus~\cite{chemrag}, and all academic benchmark corpora are available online. Frontier LLMs in our comparisons may have been pre-trained on data from these sources. The performance gaps we report are therefore not an artifact of private versus public knowledge; they demonstrate that \textit{how} a corpus is indexed, enriched, and retrieved at inference time dominates whether a model can retrieve and use knowledge it may theoretically have encountered during training. This is a more fundamental claim: retrieval quality determines accuracy for knowledge-intensive tasks even when the underlying information is publicly available.

Table~\ref{tab:retrieval_configs} characterizes the hit@10 and p95 latency of five UniK retrieval configurations across three standard benchmarks. Cross-system fusion of lexical and dense signals (\textit{UniK-Realtime}) achieves 91.2\% mean hit@10 at under 70~ms p95 latency, within 0.6 percentage points of the highest-accuracy configuration (\textit{UniK-Precision}) at 4--5$\times$ lower latency.

\begin{table}[ht]
\caption{Retrieval hit@10 and p95 latency across UniK configurations on HotpotQA (7,405 queries), NQ (2,837 queries), and PubMedQA (1,000 queries). Configurations differ in fusion strategy; all use the same enriched Polymath Index. Lexical-only and semantic-only baselines shown for reference. \textbf{Highlighted}: Best accuracy coupled with low latency. \textbf{Bold}: best result per row.}
\label{tab:retrieval_configs}
\centering
\begin{tabular}{@{}lccccr@{}}
\toprule
\textbf{Configuration} & \textbf{HotpotQA} & \textbf{NQ} & \textbf{PubMedQA} & \textbf{Mean} & \textbf{p95 lat.} \\
\midrule
Lexical only (BM25) & 91.5\% & 62.3\% & 93.0\% & 82.3\% & 61~ms \\
Semantic only (dense) & 79.9\% & 79.3\% & 93.7\% & 84.3\% & 51~ms \\
UniK-Compact & 92.4\% & 76.8\% & 96.5\% & 88.6\% & 59~ms \\
\rowcolor{blue!20} UniK-Realtime & 95.1\% & 82.7\% & 95.7\% & 91.2\% & \textbf{68~ms} \\
UniK-Precision\textsuperscript{†} & \textbf{95.9\%} & \textbf{83.4\%} & \textbf{96.2\%} & \textbf{91.8\%} & 748~ms \\
\bottomrule
\end{tabular}
\par{\footnotesize\noindent\textsuperscript{†}~Recommended for offline batch evaluation; p95 latency on NQ driven by large corpus + multi-field query expansion.}
\end{table}

\begin{table}[h!]
\caption{UniK leaderboard: performance vs.\ frontier proprietary LLMs across all five evaluation domains. UniK uses Llama-3.3-70B-Instruct (open-source, 70B parameters); frontier model sizes are undisclosed but estimated at 10--100$\times$ larger. $\dagger$~Frontier LLMs answer without access to the domain corpus (parametric memory only); UniK retrieves from the full indexed corpus at inference time. \textbf{Bold}: best result per row.}
\label{tab:leaderboard}
\centering
\setlength{\tabcolsep}{5pt}
\begin{tabular}{@{}llllcc r@{}}
\toprule
\textbf{Domain} & \textbf{Modality} & \textbf{Benchmark} & \textbf{Metric} & \textbf{UniK} & \textbf{Best Frontier$^\dagger$} & \textbf{$\Delta$} \\
\midrule
Medical       & Text       & PubMedQA        & RAG   & \textbf{77.9\%} & 71.6\%~(GPT-3.5)  & $+$6.3~pp  \\
Open Domain   & Text       & NQ              & Hit@5 & \textbf{60.5\%} & 50.0\%~(baseline) & $+$10.5~pp \\
Chemistry     & Molecular  & MolInstruct     & EM    & \textbf{64.5\%} & 53.2\%~(GPT-4o)   & $+$11.3~pp \\
Chemistry     & Molecular  & SciBench        & Score & \textbf{18.6\%} & 8.6\%~(GPT-4o)    & $+$10.0~pp \\
Legal Video   & Video      & LocalView       & RAG   & \textbf{85.7\%} & 85.7\%~(GPT-4.1)  & tied       \\
Legal Video   & Video      & Seattle CDP     & RAG   & \textbf{79.3\%} & 76.0\%~(Claude)   & $+$3.3~pp  \\
Gov.\ Data    & Structured & NJ OD (Golden)  & RAG   & \textbf{76.0\%} & 47.3\%~(GPT-5)    & $+$29.0~pp \\
Gov.\ Data    & Structured & NJ OD (Bronze)  & RAG   & \textbf{64.3\%} & 56.0\%~(Gemini)   & $+$8.3~pp  \\
\bottomrule
\end{tabular}
\end{table}

\subsection{Rich Text and Medical Intelligence}
\label{sec:text}

We evaluate on three standard open-domain and medical QA benchmarks: PubMedQA~\cite{jin2019pubmedqadataset} (62,249 biomedical documents), Natural Questions~\cite{47761} (5M documents), and SQuAD~\cite{rajpurkar2016squad} (2,067 documents, 10,570 questions). Table~\ref{tab:metagen_retrieval} shows retrieval accuracy with and without UniK enrichment. Gains are largest where the semantic gap between query and document is widest: PubMedQA improves by 7.2 percentage points (hit@1) and NQ by over 10 points (hit@5) relative to the baseline without metadata. Even on SQuAD, where the baseline is already high at 93.3\%, enrichment adds 39 correct retrievals across the development set.

\begin{table}[ht]
\caption{Impact of UniK enrichment on retrieval accuracy. Baseline uses the same index structure without enrichment metadata.}
\label{tab:metagen_retrieval}
\centering
\begin{tabular}{@{}lccc@{}}
\toprule
\textbf{Configuration} & \textbf{PubMedQA (hit@1)} & \textbf{NQ (hit@5)} & \textbf{SQuAD (hit@5)} \\
\midrule
Hybrid, no metadata & 77.3\% & 49.99\% & 93.30\% \\
+ existing metadata fields & 78.8\% & 59.49\% & 93.58\% \\
\textbf{+ UniK enrichment} & \textbf{82.1\%} & \textbf{60.48\%} & \textbf{93.68\%} \\
\midrule
Improvement over baseline & $+$4.8~pp & $+$10.5~pp & $+$0.38~pp \\
\bottomrule
\end{tabular}
\end{table}

\begin{table}[ht]
\caption{Answer accuracy on open-domain benchmarks: exact match vs.\ LLM-Judge. Exact match (EM) significantly understates true accuracy for NQ and HotpotQA because it rejects semantically-equivalent answers (e.g., ``Bill Clinton'' vs.\ ``William Jefferson Clinton''). LLM-Judge (lenient, Grade~$\geq$~2 on a 1--3 scale) recovers 20--26 percentage points by accepting correct paraphrases. PubMedQA is excluded here: its yes/no/maybe constraint makes EM equal to LLM-Judge.}
\label{tab:rag_accuracy}
\centering
\begin{tabular}{@{}lcc@{}}
\toprule
\textbf{Dataset} & \textbf{EM (strict)} & \textbf{LLM-Judge (lenient)} \\
\midrule
Natural Questions  & 41.2\% & \textbf{71.0\%} \\
HotpotQA           & 48.3\% & \textbf{74.3\%} \\
\bottomrule
\end{tabular}
\par{\footnotesize\noindent LLM-Judge grades: 3 = Excellent, 2 = Acceptable, 1 = Poor. The lenient threshold (Grade~$\geq$~2) matches human judgement for open QA; the 20--26~pp EM undercount is attributable entirely to paraphrase mismatch, not to incorrect answers.}
\end{table}

For end-to-end question answering, enriched retrieval translates directly to downstream accuracy gains. On PubMedQA, the UniK RAG pipeline achieves 77.9\% without fine-tuning, second only to RankRAG~\cite{yu2024rankrag} (79.8\%, fine-tuned) and ahead of all other non-fine-tuned methods including GPT-3.5 + RAG (71.6\%) and all other fine-tuned baselines (Table~\ref{tab:pubmedqa_rag}). The mechanism is direct: medical domain knowledge (synonyms such as ``MI'' for myocardial infarction, specialized acronyms, topical phrases) encoded in UniK enrichment fields closes the vocabulary gap that hobbles off-the-shelf embedding models in specialized domains.

\begin{table}[ht]
\caption{RAG accuracy on PubMedQA. UniK~\cite{sawarkar2025metagen,sawarkar2024blended} is the only non-fine-tuned system above 74\%.}
\label{tab:pubmedqa_rag}
\centering
\begin{tabular}{@{}lcc@{}}
\toprule
\textbf{System} & \textbf{Accuracy} & \textbf{Fine-tuned?} \\
\midrule
RankRAG~\cite{yu2024rankrag} & 79.8\% & Yes \\
\textbf{UniK (this work)} & \textbf{77.9\%} & \textbf{No} \\
AlzheimerRAG~\cite{lahiri2024alzheimerrag} & 74.0\% & Yes \\
RAFT (LLaMA2-7B)~\cite{zhang2024raft} & 73.3\% & Yes \\
GPT-3.5 + RAG~\cite{zhang2024raft} & 71.6\% & No \\
MEDRAG + GPT-4~\cite{zhao2025medrag} & 70.6\% & No \\
LLaMA2-7B + RAG~\cite{zhang2024raft} & 58.8\% & No \\
\bottomrule
\end{tabular}
\end{table}

\subsection{Chemistry and Pharmaceutical Intelligence}
\label{sec:chem}

Chemistry is among the most demanding benchmarks for universal retrieval: the ChemRAG corpus spans over 100 million documents across PubChem, PubMed, USPTO patents, Semantic Scholar, and Wikipedia~\cite{chemrag}. The benchmark tasks range from knowledge recall (MMLU-Chem) to structured prediction (ChemBench4K) to molecular generation (Mol-Instructions) to quantitative scientific reasoning (SciBench). No single retrieval strategy consistently dominates across all tasks~\cite{chemrag}, which is precisely the setting where Polymath Retrieval adds the most value.

Table~\ref{tab:chemrag} shows results from the public ChemRAG leaderboard~\cite{chemrag_leaderboard}. UniK with Llama-3.1-70B-Instruct outperforms GPT-4o with the ChemRAG paper's own retrieval baseline~\cite{chemrag} on two of four benchmarks, MolInstruct exact match (64.5\% vs. 53.2\%) and SciBench quantitative accuracy (18.6\% vs. 8.6\%), the two most demanding generative and computational tasks. On ChemBench4K, GPT-4o with domain-tuned retrieval leads (67.3\% vs. 58.6\%), while UniK with the same 70B model outperforms the identical model without UniK retrieval by more than 2$\times$ on ChemBench4K (58.6\% vs. 26.3\%) and MolInstruct (64.5\% vs. 49.7\%). The results establish UniK as the strongest open-source pipeline on ChemRAG and demonstrate competitive performance against closed proprietary systems on the hardest task categories.

\begin{table}[ht]
\caption{ChemRAG benchmark results. UniK uses Llama-3.1-70B-Instruct; all other configurations use the ChemRAG evaluation protocol. Bold: best within open-source models. ChemBench4K and MolInstruct report average accuracy and exact match (EM) respectively; SciBench reports score with tolerance. \textbf{Highlighted}: This paper. \textbf{Bold}: best result per row.}
\label{tab:chemrag}
\centering
\begin{tabular}{@{}lcccc@{}}
\toprule
\textbf{System} & \textbf{ChemBench4K} & \textbf{MolInstruct EM} & \textbf{MMLU-Chem} & \textbf{SciBench} \\
\midrule
GPT-4o + ChemRAG & \textbf{67.3\%} & 53.2\% & \textbf{73.9\%} & 8.6\% \\
o1 + ChemRAG & 58.4\% & 45.0\% & 85.5\% & 43.6\% \\
\midrule
\rowcolor{blue!20}UniK + Llama-3.1-70B & 58.6\% & \textbf{64.5\%} & 66.0\% & \textbf{18.6\%} \\
Llama-3.1-70B + ChemRAG & 26.3\% & 49.7\% & 61.1\% & 13.6\% \\
Llama-3.1-8B + ChemRAG & 25.9\% & 41.1\% & 52.2\% & 3.6\% \\
\bottomrule
\end{tabular}
\end{table}

\subsection{Legal Video Intelligence}
\label{sec:legal}

Legal proceedings, government hearings, and public testimony represent a large and largely untapped enterprise knowledge modality: video archives where the information is encoded in speech, visual context, and the structured flow of legal argument across thousands of hours of footage. We evaluate UniK on two corpora: LocalView, spanning over 1,000 hours of local government meeting video from multiple U.S. states, and Seattle City Council proceedings (Seattle CDP), spanning about 1,200 hours of council video~\cite{legalvideo_leaderboard}.

\begin{figure}[htbp]
\centering
\begin{subfigure}[t]{0.32\textwidth}
  \fbox{\includegraphics[width=\linewidth]{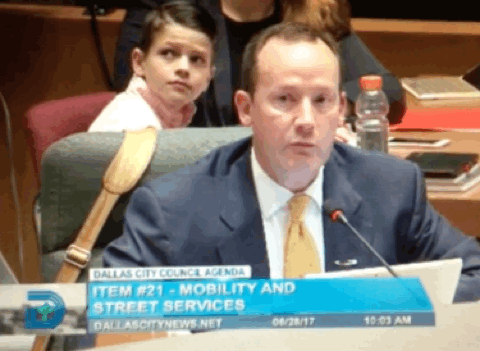}}
  \vspace{2pt}
  \par\scriptsize\itshape\sffamily Factual lookup: council vote retrieval from Seattle CDP's $\sim$1,200 hours of indexed video.
\end{subfigure}\hfill
\begin{subfigure}[t]{0.32\textwidth}
  \fbox{\includegraphics[width=\linewidth]{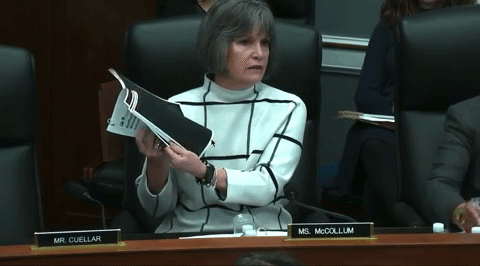}}
  \vspace{2pt}
  \par\scriptsize\itshape\sffamily Speaker attribution: identifying who said what across multi-speaker hearings.
\end{subfigure}\hfill
\begin{subfigure}[t]{0.32\textwidth}
  \fbox{\includegraphics[width=\linewidth]{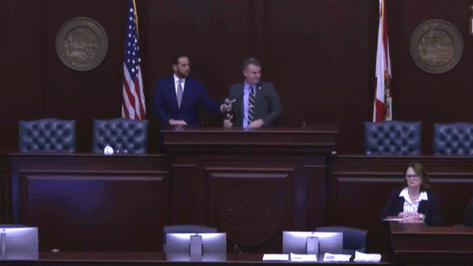}}
  \vspace{2pt}
  \par\scriptsize\itshape\sffamily Legislative context: multi-hop retrieval across sessions and amendments.
\end{subfigure}
\caption{UniK legal video intelligence in action on the Seattle City Council proceedings corpus. Answers are grounded in the indexed transcript, not recalled from model weights.}
\label{fig:legal_demo}
\end{figure}

The UniK video pipeline extends the core Polymath Retrieval architecture with multimodal enrichment: Vision Language Model scene understanding and prosodic audio analysis augment transcript-based indices, while an automatically constructed knowledge graph enables structured entity-relationship queries across the corpus. Results on curated golden question sets (50 questions per corpus) and large bronze sets (928--1,000 production queries) are shown in Tables~\ref{tab:legal_golden} and~\ref{tab:legal_bronze}. Figure~\ref{fig:legal_demo} shows the UniK legal video intelligence interface answering production queries against the Seattle CDP corpus; each panel captures a different query type (factual lookup, speaker attribution, and multi-turn legislative context).

\begin{table}[ht]
\caption{Legal video benchmark -- Golden set (50 curated questions per corpus). Avg Grade on a 1--3 scale (3 = Excellent, 1 = Poor; higher is better). RAG Accuracy = avg\_grade / 3 $\times$ 100. Frontier LLMs answer without access to the video corpus; UniK retrieves from LocalView (1,000+ hours) and Seattle CDP ($\sim$1,200 hours) using Llama-3.3-70B-Instruct.}
\label{tab:legal_golden}
\centering
\begin{tabular}{@{}lcccc@{}}
\toprule
\multirow{2}{*}{\textbf{System}} & \multicolumn{2}{c}{\textbf{RAG Accuracy $\uparrow$}} & \multicolumn{2}{c}{\textbf{Avg Grade (3 = best) $\uparrow$}} \\
\cmidrule(lr){2-3} \cmidrule(lr){4-5}
 & LocalView & Seattle CDP & LocalView & Seattle CDP \\
\midrule
\textbf{UniK + Llama-3.3-70B} & 77.7\% & \textbf{82.0\%} & 2.33 & \textbf{2.46} \\
GPT-4.1 & \textbf{81.3\%} & 74.7\% & \textbf{2.44} & 2.24 \\
Claude Sonnet 4.6 & 64.3\% & 71.3\% & 1.93 & 2.14 \\
Gemini 2.0 Flash & 63.3\% & 62.7\% & 1.90 & 1.88 \\
\bottomrule
\end{tabular}
\end{table}

\begin{table}[ht]
\caption{Legal video benchmark -- Bronze set (928--1{,}000 production queries per corpus). Same grading scale and RAG Accuracy formula as Table~\ref{tab:legal_golden}.}
\label{tab:legal_bronze}
\centering
\begin{tabular}{@{}lcccc@{}}
\toprule
\multirow{2}{*}{\textbf{System}} & \multicolumn{2}{c}{\textbf{RAG Accuracy $\uparrow$}} & \multicolumn{2}{c}{\textbf{Avg Grade (3 = best) $\uparrow$}} \\
\cmidrule(lr){2-3} \cmidrule(lr){4-5}
 & LocalView & Seattle CDP & LocalView & Seattle CDP \\
\midrule
\textbf{UniK + Llama-3.3-70B} & \textbf{85.7\%} & \textbf{79.3\%} & \textbf{2.57} & \textbf{2.38} \\
GPT-4.1 & \textbf{85.7\%} & 73.7\% & \textbf{2.57} & 2.21 \\
Claude Sonnet 4.6 & 73.3\% & 76.0\% & 2.20 & 2.28 \\
Gemini 2.0 Flash & 64.7\% & 67.0\% & 1.94 & 2.01 \\
\bottomrule
\end{tabular}
\end{table}

UniK leads on Seattle CDP across both evaluation sets (2.46 Golden, 2.38 Bronze), the corpus where retrieval from $\sim$1,200 hours of indexed video matters most. On LocalView, GPT-4.1 edges ahead on the Golden set (2.44 vs.\ 2.33) and ties at the Bronze scale (2.57 vs.\ 2.57), while UniK uses a model roughly 20$\times$ smaller. All frontier LLMs answer without access to the combined 2,200+ hours of indexed video across both corpora; they rely on parametric memory or limited context, while UniK retrieves and grounds answers in the actual proceedings. This comparison is not model against model but a retrieval-augmented open-source model against a massive proprietary model operating from memory.

\subsection{Government and Open Data Intelligence}
\label{sec:gov}

We evaluate on the New Jersey State Open Data corpus~\cite{sawarkar2025metagen}, a real-world enterprise knowledge base covering state government datasets spanning public health, transportation, economics, and environmental domains across text and tabular modalities. This benchmark isolates the challenge of retrieval-grounded accuracy: answering questions about specific data points, legislation, or agency decisions requires finding the relevant record in a large heterogeneous corpus, not inferring from general world knowledge.

Table~\ref{tab:njod} shows results from the public leaderboard~\cite{njod_leaderboard}. UniK achieves 76\% RAG accuracy on the Golden set versus 47\% for GPT-5, 46\% for Gemini-3 Flash, and 42\% for Claude Sonnet 4.6. The pattern is consistent across the Bronze set (635 queries), where UniK maintains an 8-percentage-point lead over the next best frontier model.

\begin{table}[ht]
\caption{NJ Open Data benchmark. RAG Accuracy = avg\_grade / 3 $\times$ 100; Grade 3 = excellent, Grade 1 = poor (higher average grade = better). All frontier LLMs answer without corpus retrieval.}
\label{tab:njod}
\centering
\begin{tabular}{@{}lcccc@{}}
\toprule
\multirow{2}{*}{\textbf{System}} & \multicolumn{2}{c}{\textbf{RAG Accuracy $\uparrow$}} & \multicolumn{2}{c}{\textbf{Avg Grade (3 = best) $\uparrow$}} \\
\cmidrule(lr){2-3} \cmidrule(lr){4-5}
 & Golden & Bronze & Golden & Bronze \\
\midrule
\textbf{UniK + Llama-3.3-70B} & \textbf{76.0\%} & \textbf{64.3\%} & \textbf{2.28} & \textbf{1.93} \\
GPT-5 & 47.3\% & 52.3\% & 1.42 & 1.57 \\
Gemini-3 Flash & 46.0\% & 56.0\% & 1.38 & 1.68 \\
Claude Sonnet 4.6 & 42.0\% & 39.7\% & 1.26 & 1.19 \\
\bottomrule
\end{tabular}
\end{table}

The NJ Open Data results make the retrieval argument concrete. Even if an LLM has encountered snapshots of this public data during pretraining, it cannot reliably cite the specific regulatory table, the exact agency decision by date, or the current numeric value of a government indicator; parametric knowledge is stale, compressed, and unverifiable. Retrieval from an indexed, enriched corpus gives UniK a 29-percentage-point accuracy advantage on the Golden set precisely because the answer is grounded in the source document rather than recalled from training weights. UniK's 76\% accuracy reflects current pipeline coverage on a heterogeneous corpus with inconsistent semi-structured formatting; the accuracy ceiling is retrieval coverage, not model capability.

\section{The Efficiency Argument}

Across all five domains, the information bottleneck is retrieval quality rather than model scale. Table~\ref{tab:leaderboard} summarizes performance of UniK + Llama-3.3-70B against leading frontier LLMs across all domains and modalities. In every domain where grounded knowledge is essential (government data, specialized medical QA, chemical generation, legal video at scale) a 70B open-source model with strong retrieval matches or exceeds models that are an order of magnitude larger and accessed exclusively via proprietary APIs.

Enterprises do not need the largest or most expensive proprietary models to achieve frontier-level performance on knowledge-intensive tasks. Investing in the retrieval and enrichment layer makes a smaller open-source model competitive, with the added benefits of cost efficiency, data privacy, and on-premise deployment.

The efficiency gap is largest precisely where the knowledge is most private and domain-specific: NJ government data, biomedical literature, chemical synthesis. In these settings, a larger model's richer parametric knowledge is irrelevant because the knowledge required was never in its training data. What matters is whether the retrieval layer can surface the right document from the enterprise corpus. This is the argument for investing in universal knowledge perception infrastructure rather than chasing larger models.

\section{Universal Knowledge Perception for Physical AI}

Physical AI (robots, autonomous vehicles, and embodied agents that must act in the real world) requires knowledge infrastructure that is fundamentally different in modality but structurally identical in operation to what digital AI requires. This section makes the case that UniK is the right platform for both, and that the UniK platform provides a concrete path from the text-based results of the preceding sections to the sensor- and video-based data challenges of physical AI. The benchmark results in Sections~\ref{sec:text}--\ref{sec:gov} demonstrate what UniK achieves for digital AI; the architecture and use cases below describe how the same platform addresses the analogous problem for physical AI.

\subsection{The Knowledge Problem in Physical AI}

Physical AI systems (robots, autonomous vehicles, embodied agents) learn by training on large corpora of interaction data: video, gameplay demonstrations, egocentric observations, and sensor telemetry. The quality of these systems depends not only on the architecture of the world model or policy but critically on the quality of the data pipeline that curates the training corpus. This pipeline faces the same knowledge perception challenge that digital AI does, at larger scale and under harder constraints---physical validity, coverage-aware sampling, and real-time ingestion among them, enumerated in full in Section~\ref{sec:open_challenges}.

The unifying architecture for physical AI is the \textit{world model}: an internal simulator trained to predict how the environment evolves given observations and actions. World models enable sample-efficient policy learning and planning without costly real-world rollouts~\cite{survey_world_models_2025, survey_embodied_2025}. Leading examples include Meta's V-JEPA 2~\cite{vjepa2_2025}, trained on over one million hours of internet video to achieve zero-shot robot manipulation on unseen hardware, and NVIDIA's GR00T N1~\cite{gr00t_n1_2025}, a humanoid foundation model combining a vision-language planner with a diffusion-based motor action model. LeCun's Joint Embedding Predictive Architecture (JEPA)~\cite{lecun2022jepa} provides a theoretical grounding for this direction: rather than predicting raw pixels, world models should predict in abstract latent space, learning representations of dynamics and structure that generalize across embodiments.

The data bottleneck is underappreciated relative to the architectural progress receiving most of the attention. A world model for robotics needs training episodes that are semantically relevant (similar tasks and object configurations), physically valid (kinematically consistent trajectories), distribution-covering (diverse enough to prevent mode collapse in learned dynamics), and appropriately sequenced for curriculum learning. None of these properties can be specified by hand at the scale of millions of episodes. They require automatic enrichment, Polymath Indexing over multiple signal types, and efficient Polymath Retrieval under task-specific queries: exactly the operations UniK performs for digital AI.

\subsection{The Common Infrastructure}

The UniK architecture applies directly to physical AI data pipelines. Table~\ref{tab:UniK_comparison} shows how the six UniK operations map across digital and physical AI use cases.

\begin{table}[ht]
\caption{UniK operations are common across digital and physical AI; modality-specific enrichment is the principal point of differentiation.}
\label{tab:UniK_comparison}
\centering
\begin{tabular}{@{}p{2.4cm}p{4.8cm}p{4.8cm}@{}}
\toprule
\textbf{UniK Operation} & \textbf{Digital AI} & \textbf{Physical AI} \\
\midrule
Ingestion & Batch documents, streaming articles, video transcripts & Telemetry streams, video episodes, gameplay recordings \\
Enrichment & Keyphrases, named entities, synonyms, topics & Object labels, action types, kinematic features, failure flags \\
Indexing & BM25 lexical + dense semantic, metadata-boosted & Spatial + temporal + visual + kinematic hybrid indices \\
Retrieval & Polymath Retrieval (RRF fusion over text, vector, and graph indices) & Episode retrieval by behavior specification or failure pattern \\
Evaluation & Retrieval accuracy on labeled QA benchmarks & Curriculum coverage, training distribution metrics \\
Continuous learning & Incremental index updates as documents change & Online ingestion of new fleet episodes and demonstrations \\
\bottomrule
\end{tabular}
\end{table}

Both digital and physical AI face the same root problem: the knowledge an AI system needs to function is distributed across a large, heterogeneous, evolving corpus that cannot be fully loaded into any model's context or training batch. The solution in both cases is a retrieval infrastructure that enriches items with metadata, indexes them for efficient lookup, and retrieves the most relevant subset on demand. UniK is that infrastructure.

\subsection{Physical AI Use Cases}

Three use cases show how UniK applies to the physical AI data pipeline:

\textbf{UC1: Paired Video and Action Data (Gameplay / Teleoperation).} Robot learning from human demonstrations requires pairing video observations with action streams (joint angles, end-effector poses, controller inputs). A semantic retrieval layer must align video and action streams temporally, validate physical consistency of trajectories, and sample training batches that maximize coverage of the task distribution. Automated curriculum generation (surfacing progressively harder episodes as the policy improves) is a direct application of the retrieval and evaluation capabilities in UniK.

\textbf{UC2: Egocentric Video Without Action Labels.} Large quantities of egocentric video (first-person human demonstrations, wearable cameras) exist without corresponding action labels. Pretraining visual encoders and bootstrapping latent action spaces from this data requires solving an \textit{inverse dynamics} problem: inferring the implicit action or state transition that connects consecutive observations, without ever observing the action signal directly. This differs from conventional video retrieval (finding a clip by keyword): it requires identifying clips with meaningful, learnable state transitions---hand-object contact, manipulation events, navigational transitions---and demands multimodal enrichment (object detection, hand pose estimation, scene classification) combined with temporal index structures that track state change across a sequence, not just per-frame semantic similarity.

\textbf{UC3: Robot Fleet Telemetry.} Deployed robot fleets generate continuous streams of high-frequency sensor data: joint torques, force-torque readings, IMU signals, camera feeds, error logs. Operational intelligence (identifying failure modes, correlating failures with environmental conditions, retrieving similar past incidents) requires a retrieval system that can embed temporal signal episodes, detect anomalies, cluster failure signatures, and surface relevant historical data in response to a new failure event.

\subsection{Open Challenges}
\label{sec:open_challenges}

UniK for physical AI faces challenges beyond the digital case:

\begin{itemize}
  \item \textbf{Multi-modal temporal alignment:} video, action streams, and sensor telemetry must be aligned to sub-second precision before indexing; misalignment corrupts the semantic content of the episode.
  \item \textbf{Physical validity:} unlike text, where correctness is semantic, physical training data must satisfy kinematic and dynamic constraints; invalid trajectories can harm policy learning even if semantically similar to valid ones.
  \item \textbf{Coverage-aware sampling:} world model training is highly sensitive to the distribution of training data; retrieval must be diversity-aware, not just relevance-aware, to avoid mode collapse in the learned dynamics.
  \item \textbf{Scale and real-time ingestion:} deployed robot fleets produce data at rates that require streaming ingestion, online anomaly detection, and near-real-time indexing, which go beyond the batch processing assumed by most RAG architectures.
  \item \textbf{Evaluation:} unlike text QA, there is no simple ground truth for whether a retrieved training batch will improve a world model; evaluation requires a closed loop with model training, making benchmark construction for physical AI retrieval significantly harder.
\end{itemize}

Despite these open problems, the structural case for UniK as the foundation of physical AI is strong. The operations are the same as for digital AI; the engineering adapters differ. Organizations already using UniK for enterprise document retrieval can extend the same infrastructure to physical AI data curation with modality-specific enrichment as the adaptation layer rather than building a separate data pipeline from scratch. The lesson from digital AI transfers directly: investing in the knowledge infrastructure layer compounds across every downstream use case, and physical AI is the next domain where that investment will prove decisive.

\section{Conclusion}

Digital AI and physical AI are converging on the same foundational bottleneck: the knowledge they need to function is heterogeneous, domain-specific, multi-modal, and distributed across corpora that are too large, too private, and too dynamic for any model to internalize through training alone. Universal Knowledge Perception is the paradigm we propose to address this bottleneck across both classes, not as two separate efforts but as a single platform with shared operations and modality-specific adapters.

UniK validates this claim on the digital AI side. Across five domains (medical, open-domain, chemistry, legal video, and government data) domain-agnostic Polymath Retrieval combined with an open-source 70B-parameter model consistently matches or outperforms frontier proprietary models that are estimated to be orders of magnitude larger. The central finding is that retrieval quality matters more than model scale for grounded, knowledge-intensive tasks: investing in the knowledge infrastructure layer compounds across every domain and use case simultaneously.

The physical AI extension makes the platform argument concrete. World model training, visual encoder pretraining from egocentric video, and robot fleet intelligence are all data curation and retrieval problems. The same six UniK operations (ingestion, enrichment, indexing, retrieval, evaluation, and continuous learning) apply directly, with modality-specific enrichment as the principal adaptation. The architectural lesson from digital AI transfers: no single retrieval method dominates, Polymath Retrieval over enriched indices is the right default, and domain generality requires investing in the platform rather than in per-task adapters.

\textbf{Open directions.} Several challenges remain open for extending Universal Knowledge Perception fully into the physical and virtual domains. Auto-didactic retrieval (Section~\ref{sec:platform})---learning per-domain retrieval configuration and continuously adapting it online---remains unsolved outside of narrow hyperparameter tuning. On the physical AI side, multi-modal temporal alignment, physical validity checking, and coverage-aware sampling (Section~\ref{sec:open_challenges}) require enrichment and indexing techniques with no direct digital AI analogue, and evaluation itself is an open problem: unlike text QA, there is no simple ground truth for whether a retrieved training batch improves a world model. Extending UniK to naturalistic, non-verbal video (bodycam and surveillance streams, where the dominant signal is visual rather than linguistic) and to action-rich virtual environments (game and simulation data reuse for physical AI training) are the two directions we view as most immediately tractable, and where we intend to focus future work.

We view UniK as infrastructure in the same sense that databases were infrastructure for transactional computing: a layer that every AI application will depend on, that rewards investment because the benefits compound across use cases, and that will look obvious in retrospect. UniK is our implementation of that infrastructure, and this paper is the first account of its performance across the breadth of domains it is designed to serve.

\bibliographystyle{plain}
\bibliography{references}

@misc{barnett2024sevenfailurepoints,
      title={Seven Failure Points When Engineering a Retrieval Augmented Generation System}, 
      author={Scott Barnett and Stefanus Kurniawan and Srikanth Thudumu and Zach Brannelly and Mohamed Abdelrazek},
      year={2024},
      eprint={2401.05856},
      archivePrefix={arXiv},
      primaryClass={cs.SE},
      url={https://arxiv.org/abs/2401.05856}, 
}

@misc{du2025contextlengthhurtsllm,
      title={Context Length Alone Hurts LLM Performance Despite Perfect Retrieval}, 
      author={Yufeng Du and Minyang Tian and Srikanth Ronanki and Subendhu Rongali and Sravan Bodapati and Aram Galstyan and Azton Wells and Roy Schwartz and Eliu A Huerta and Hao Peng},
      year={2025},
      eprint={2510.05381},
      archivePrefix={arXiv},
      primaryClass={cs.CL},
      url={https://arxiv.org/abs/2510.05381}, 
}

@misc{hsieh2024rulerwhatsrealcontext,
      title={RULER: What's the Real Context Size of Your Long-Context Language Models?}, 
      author={Cheng-Ping Hsieh and Simeng Sun and Samuel Kriman and Shantanu Acharya and Dima Rekesh and Fei Jia and Yang Zhang and Boris Ginsburg},
      year={2024},
      eprint={2404.06654},
      archivePrefix={arXiv},
      primaryClass={cs.CL},
      url={https://arxiv.org/abs/2404.06654}, 
}

@misc{jin2019pubmedqadataset,
      title={PubMedQA: A Dataset for Biomedical Research Question Answering}, 
      author={Qiao Jin and Bhuwan Dhingra and Zhengping Liu and William W. Cohen and Xinghua Lu},
      year={2019},
      eprint={1909.06146},
      archivePrefix={arXiv},
      primaryClass={cs.CL},
      url={https://arxiv.org/abs/1909.06146}, 
}

@techreport{mit2025genaidivide,
  author      = "{MIT Project NANDA}",
  title       = "The GenAI Divide: State of AI in Business 2025",
  institution = "Massachusetts Institute of Technology",
  year        = "2025",
  month       = "July",
  note        = "Found in: https://www.legal.io/articles/5719519/MIT-Report-Finds-95-of-AI-Pilots-Fail-to-Deliver-ROI-Exposing-GenAI-Divide"
}

@misc{reimers2021cursedenselowdimensional,
      title={The Curse of Dense Low-Dimensional Information Retrieval for Large Index Sizes}, 
      author={Nils Reimers and Iryna Gurevych},
      year={2021},
      eprint={2012.14210},
      archivePrefix={arXiv},
      primaryClass={cs.IR},
      url={https://arxiv.org/abs/2012.14210}, 
}

@inproceedings{sawarkar2024blended,
  title={Blended RAG: Improving RAG (Retriever-Augmented Generation) Accuracy with Semantic Search and Hybrid Query-Based Retrievers},
  author={Sawarkar, Kunal and Mangal, Abhilasha and Solanki, Shivam Raj},
  booktitle={The 7th IEEE International Conference on Multimedia Information Processing and Retrieval (IEEE-MIPR 2024)},
  year={2024},
  publisher={IEEE},
  doi={10.1109/MIPR62202.2024.00031},
  url={https://doi.org/10.48550/arXiv.2404.07220}
}

@misc{sawarkar2025metagen,
  title={MetaGen Blended RAG: Unlocking Zero-Shot Precision for Specialized Domain Question-Answering},
  author={Sawarkar, Kunal and Solanki, Shivam R. and Mangal, Abhilasha},
  year={2025},
  url={https://doi.org/10.48550/arXiv.2505.18247},
  eprint={2505.18247},
  archivePrefix={arXiv},
  primaryClass={cs.CL}
}

@article{siriwardhana-etal-2023-improving,
    title = "Improving the Domain Adaptation of Retrieval Augmented Generation ({RAG}) Models for Open Domain Question Answering",
    author = "Siriwardhana, Shamane  and
      Weerasekera, Rivindu  and
      Wen, Elliott  and
      Kaluarachchi, Tharindu  and
      Rana, Rajib  and
      Nanayakkara, Suranga",
    journal = "Transactions of the Association for Computational Linguistics",
    volume = "11",
    year = "2023",
    address = "Cambridge, MA",
    publisher = "MIT Press",
    url = "https://aclanthology.org/2023.tacl-1.1/",
    doi = "10.1162/tacl_a_00530",
    pages = "1--17"
}

@misc{weller2026-embeddingbasedretrieval,
      title={On the Theoretical Limitations of Embedding-Based Retrieval}, 
      author={Orion Weller and Michael Boratko and Iftekhar Naim and Jinhyuk Lee},
      year={2026},
      eprint={2508.21038},
      archivePrefix={arXiv},
      primaryClass={cs.IR},
      url={https://arxiv.org/abs/2508.21038}, 
}

@misc{zhao2025medrag,
      title={MedRAG: Enhancing Retrieval-augmented Generation with Knowledge Graph-Elicited Reasoning for Healthcare Copilot}, 
      author={Xuejiao Zhao and Siyan Liu and Su-Yin Yang and Chunyan Miao},
      year={2025},
      eprint={2502.04413},
      archivePrefix={arXiv},
      primaryClass={cs.CL},
      url={https://arxiv.org/abs/2502.04413}, 
}

@article{rajpurkar2016squad,
  title={SQuAD: 100,000+ questions for machine comprehension of text},
  author={Rajpurkar, Pranav and Zhang, Jian and Lopyrev, Konstantin and Liang, Percy},
  journal={arXiv preprint arXiv:1606.05250},
  year={2016}
}

@inproceedings{zhang2024raft,
  title={RAFT: Adapting Language Model to Domain Specific RAG},
  author={Zhang, Tianjun and Patil, Shishir G and Jain, Naman and Shen, Sheng and Zaharia, Matei and Stoica, Ion and Gonzalez, Joseph E},
  booktitle={First Conference on Language Modeling},
  year={2024}
}

@article{lahiri2024alzheimerrag,
  title={AlzheimerRAG: Multimodal Retrieval Augmented Generation for PubMed articles},
  author={Lahiri, Aritra Kumar and Hu, Qinmin Vivian},
  journal={arXiv preprint arXiv:2412.16701},
  year={2024}
}

@article{yu2024rankrag,
  title={RankRAG: Unifying context ranking with retrieval-augmented generation in LLMs},
  author={Yu, Yue and Ping, Wei and Liu, Zihan and Wang, Boxin and You, Jiaxuan and Zhang, Chao and Shoeybi, Mohammad and Catanzaro, Bryan},
  journal={Advances in Neural Information Processing Systems},
  volume={37},
  pages={121156--121184},
  year={2024}
}

@article{grattafiori2024llama,
  title={The llama 3 herd of models},
  author={Grattafiori, Aaron and Dubey, Abhimanyu and Jauhri, Abhinav and Pandey, Abhinav and Kadian, Abhishek and Al-Dahle, Ahmad and Letman, Aiesha and Mathur, Akhil and Schelten, Alan and Vaughan, Alex and others},
  journal={arXiv preprint arXiv:2407.21783},
  year={2024}
}

@article{yang2018hotpotqa,
  title={HotpotQA: A dataset for diverse, explainable multi-hop question answering},
  author={Yang, Zhilin and Qi, Peng and Zhang, Saizheng and Bengio, Yoshua and Cohen, William W and Salakhutdinov, Ruslan and Manning, Christopher D},
  journal={arXiv preprint arXiv:1809.09600},
  year={2018}
}

@article{47761,title	= {Natural Questions: a Benchmark for Question Answering Research},author	= {Tom Kwiatkowski and Jennimaria Palomaki and Olivia Redfield and Michael Collins and Ankur Parikh and Chris Alberti and Danielle Epstein and Illia Polosukhin and Matthew Kelcey and Jacob Devlin and Kenton Lee and Kristina N. Toutanova and Llion Jones and Ming-Wei Chang and Andrew Dai and Jakob Uszkoreit and Quoc Le and Slav Petrov},year	= {2019},journal	= {Transactions of the Association of Computational Linguistics}}

@article{chemrag,
  title={Benchmarking Retrieval-Augmented Generation for Chemistry},
  author={Zhong, Xianrui and Jin, Bowen and Ouyang, Siru and Shen, Yanzhen and Jin, Qiao and Fang, Yin and Lu, Zhiyong and Han, Jiawei},
  journal={arXiv preprint arXiv:2505.07671},
  year={2025}
}

@misc{vjepa2_2025,
  title={V-JEPA 2: Self-Supervised Video Models Enable Understanding, Prediction and Planning},
  author={Assran, Mahmoud and Duval, Quentin and Balestriero, Randall and Misra, Ishan and Bojanowski, Piotr and Vincent, Pascal and Rabbat, Michael and LeCun, Yann and Ballas, Nicolas},
  year={2025},
  eprint={2506.09985},
  archivePrefix={arXiv},
  primaryClass={cs.CV},
  url={https://arxiv.org/abs/2506.09985}
}

@misc{gr00t_n1_2025,
  title={NVIDIA Isaac GR00T N1: An Open Foundation Model for Generalist Humanoid Robots},
  author={NVIDIA},
  year={2025},
  eprint={2503.14734},
  archivePrefix={arXiv},
  primaryClass={cs.RO},
  url={https://arxiv.org/abs/2503.14734}
}

@misc{survey_world_models_2025,
  title={A Comprehensive Survey on World Models for Embodied {AI}},
  author={Zeng, Fanqi and Liang, Bencheng and Shi, Jianhua and Shen, Wei},
  year={2025},
  eprint={2510.16732},
  archivePrefix={arXiv},
  primaryClass={cs.AI},
  url={https://arxiv.org/abs/2510.16732}
}

@misc{survey_embodied_2025,
  title={A Survey: Learning Embodied Intelligence from Physical Simulators and World Models},
  author={Zhou, Xiaoyuan and Xue, Haoyuan and Gao, Yunbiao and Huang, Jiale and Feng, Chen and Gao, Shangzhe and Cheng, Yingzi and Luo, Lin and Pan, Jiajun and Liao, Zhengwen},
  year={2025},
  eprint={2507.00917},
  archivePrefix={arXiv},
  primaryClass={cs.RO},
  url={https://arxiv.org/abs/2507.00917}
}

@misc{lecun2022jepa,
  title={A Path Towards Autonomous Machine Intelligence},
  author={LeCun, Yann},
  year={2022},
  url={https://openreview.net/pdf?id=BZ5a1r-kVsf},
  note={Open Review}
}

@article{radovanovic2010hubs,
  title={Hubs in space: Popular nearest neighbors in high-dimensional data},
  author={Radovanovi{\'c}, Milo{\v{s}} and Nanopoulos, Alexandros and Ivanovi{\'c}, Miroslav},
  journal={Journal of Machine Learning Research},
  volume={11},
  pages={2487--2531},
  year={2010}
}

@inproceedings{thakur2021beir,
  title={{BEIR}: A heterogeneous benchmark for zero-shot evaluation of information retrieval models},
  author={Thakur, Nandan and Reimers, Nils and R{\"u}ckl{\'e}, Andreas and Srivastava, Abhimanyu and Gurevych, Iryna},
  booktitle={Thirty-fifth Conference on Neural Information Processing Systems Datasets and Benchmarks Track},
  year={2021}
}

@misc{chemrag_leaderboard,
  title={{ChemRAG} Leaderboard v1},
  author={{AIntropy AI}},
  year={2025},
  howpublished={\url{https://huggingface.co/spaces/aintropy-ai/chemRAG-leaderboard-v1}}
}

@misc{legalvideo_leaderboard,
  title={Legal Videos {QA} Leaderboard v1},
  author={{AIntropy AI}},
  year={2025},
  howpublished={\url{https://huggingface.co/spaces/aintropy-ai/legal-videos-leaderboard-v1}}
}

@misc{njod_leaderboard,
  title={{NJ} Open Data {QA} Leaderboard v1},
  author={{AIntropy AI}},
  year={2025},
  howpublished={\url{https://huggingface.co/spaces/aintropy-ai/nj-open-data-leaderboard-v1}}
}

@inproceedings{autoragHP,
  title={{AutoRAG-HP}: Automatic Online Hyper-Parameter Tuning for Retrieval-Augmented Generation},
  author={Ruan, Sherry and others},
  booktitle={Findings of the Association for Computational Linguistics: EMNLP 2024},
  year={2024},
  url={https://arxiv.org/abs/2406.19251}
}

@misc{autorag2024,
  title={{AutoRAG}: Automated Framework for Optimization of Retrieval Augmented Generation Pipeline},
  author={Kim, Dongkyu and Kim, Byoungwook and Han, Donggeon and Eibich, Matouš},
  year={2024},
  eprint={2410.20878},
  archivePrefix={arXiv},
  url={https://arxiv.org/abs/2410.20878}
}

\end{document}